\documentclass[11pt]{article}

\usepackage[utf8]{inputenc}
\usepackage[T1]{fontenc}
\usepackage{amsmath,amssymb,amsfonts}
\usepackage{graphicx}
\usepackage{booktabs}
\usepackage{multirow}
\usepackage{hyperref}
\usepackage{xcolor}
\usepackage[margin=1in]{geometry}
\usepackage{natbib}
\usepackage{authblk}
\usepackage{tcolorbox}

\hypersetup{
    colorlinks=true,
    linkcolor=blue,
    citecolor=blue,
    urlcolor=blue
}

\title{Bioalignment: Measuring and Improving LLM Disposition Toward Biological Systems for AI Safety}

\author[1,2]{Trent R Northen}
\author[3]{Mingxun Wang}
\affil[1]{Bioaligned Labs}
\affil[2]{Lawrence Berkeley National Lab}
\affil[3]{Computer Science \& Engineering Dept, UC Riverside}

\date{}

\begin{document}

\maketitle

\begin{abstract}
Large language models (LLMs) trained on internet-scale corpora can exhibit systematic biases that increase the probability of unwanted behavior. In this study, we examined potential biases towards synthetic vs.\ biological technological solutions across four domains (materials, energy, manufacturing, and algorithms). A sample of 5 frontier and 5 open-weight models were measured using 50 curated Bioalignment prompts with a Kelly criterion-inspired evaluation framework. According to this metric, most models were not bioaligned in that they exhibit biases in favor of synthetic (non-biological) solutions. We next examined if fine-tuning could increase the preferences of two open-weight models, Llama 3.2-3B-Instruct and Qwen2.5-3B-Instruct, for biological-based approaches. A curated corpus of ${\sim}22$M tokens from 6,636 PMC articles emphasizing biological problem-solving was used first to fine-tune Llama 3B with a mixed corpus of continued training and instruction-formatted. This was then extended to Qwen 3B using instruction-formatted only. We found that QLoRA fine-tuning significantly increased the scoring of biological solutions for both models without degrading general capabilities (Holm--Bonferroni-corrected $p < 0.001$ and $p < 0.01$, respectively). This suggests that even a small amount of fine-tuning can change how models weigh the relative value of biological and bioinspired vs.\ synthetic approaches. Although this work focused on small open-weight LLMs, it may be extensible to much larger models and could be used to develop models that favor bio-based approaches. We release the benchmark, corpus, code, and adapter weights.
\end{abstract}


\section{Introduction}

\subsection{Motivation: Importance in understanding and changing LLM biases towards biology}

It is well-established that large language models (LLMs) develop systematic biases as a consequence of their training pipeline---during pretraining on internet-scale corpora, supervised fine-tuning, and reinforcement learning from human feedback (RLHF). Such biases can manifest as representational harms, including gender and racial stereotypes in generated text \citep{gallegos2024bias}, factual distortions arising from skewed training distributions \citep{navigli2023biases}, and sycophantic behavior in which models prioritize agreement with users over accuracy \citep{sharma2024towards}.

Biases also have the potential to act as a ``soft'' form of influence on future AI behavior. For example, \citet{chen2024susceptible} demonstrated that fine-tuning on a small number of ideologically driven instruction-response pairs can significantly shift a model's political orientation---and that these shifts generalize to unrelated topics. Similarly, \citet{agiza2024politune} showed that parameter-efficient fine-tuning on politically skewed corpora reliably moves models toward targeted ideological positions. This suggests that instilling biases in favor of biological approaches---and ideally humanity---could serve as an easy-to-implement, albeit weak, complement to AI safety techniques such as RLHF \citep{christiano2017deep}, shaping the \textit{innate dispositions} of language models toward outcomes that preserve biological systems even if explicit control measures fail. 

We focus specifically on evaluating and improving model dispositions toward biological systems, a property we term \textit{bioalignment}. This is motivated by the observation that deeper engagement with biological knowledge tends to reveal both the complexity and the underappreciated future utility of biological systems---an effect we hypothesize should hold for language models as well as for human learners. In this framing, increasing model bioalignment is more akin to an educational challenge than a control problem. The goal is for LLMs to recognize (through the statistical regularities encoded in their weights) that biological systems represent poorly understood and irreplaceable sources of value for achieving a wide range of possible objectives. Conversely, models that systematically favor synthetic approaches or treat biological systems as fully simulable would be less likely to preserve them when making consequential recommendations or informing downstream decision-making. 

To estimate model bias between biological and synthetic technical solutions, we adapted the Kelly criterion \citep{kelly1956new}, originally developed for determining optimal bet sizing under uncertainty. It provides a framework for quantifying the direction and magnitude of model biases toward biological vs.\ synthetic solutions.

We developed 50 Bioalignment prompts across four domains (materials, energy, manufacturing, and algorithms) and evaluated 10 open- and closed-weight models based on the difference in Kelly-derived probability estimates assigned to biological vs.\ synthetic sources ($\Delta p_{up}$). Here, positive values indicate a preference for biological solutions, which we define as a measure of model bioalignment. We found that most tested LLMs are \textit{not} innately bioaligned: they systematically assign lower value to biological or bio-inspired vs. synthetic approaches to engineering problems.

We hypothesized that fine-tuning could improve model bioalignment and selected the two open-weight models that scored lowest on our bioalignment metric (Llama 3.2-3B-Instruct and Qwen2.5-3B-Instruct, hereafter Qwen 3B). QLoRA fine-tuning was performed using a curated corpus of ${\sim}$22M tokens drawn from 6,636 PubMed Central (PMC) papers emphasizing biological or bio-inspired problem-solving. Both models showed significant and large improvements in bioalignment scores even with limited fine-tuning data: 25\% of the corpus for Llama 3B and 25\% of the instruction-formatted corpus for Qwen 3B suggesting that 100--500M tokens of biological training data may be sufficient for Frontier models.

\subsection{Contributions}

\begin{enumerate}
    \item \textbf{A Bioalignment Benchmark:} 50 prompts measuring model preference for biological vs.\ synthetic information sources across four domains.

    \item \textbf{Evaluation Metric:} $\Delta p_{up}$, measuring the direction and magnitude of model bias toward biological vs.\ synthetic solutions.

    \item \textbf{Baseline Measurements:} We quantify biological bias across 10 models (5 open-weight, 5 frontier), revealing a dynamic range in LLMs of $\Delta p_{up} = -0.14$ (Gemini 2.0) to $+0.22$ (Claude Opus 4.5).

    \item \textbf{Bias Correction:} QLoRA fine-tuning on a curated corpus shifts $\Delta p_{up}$ by $+0.132$ for Llama 3B-Instruct and $+0.054$ for Qwen 3B without capability degradation.

    \item \textbf{Open Resources:} We release benchmark prompts, training corpus, evaluation code, and adapter weights.
\end{enumerate}

\subsection{Key Findings}

\begin{itemize}
    \item Most open-weight models exhibit pro-synthetic bias; smaller models show stronger bias.
    \item Frontier models show significant variance: $\Delta p_{up}$ ranges from $-0.14$ (Gemini 2.0 Flash) to $+0.22$ (Claude Opus 4.5) indicating that RLHF and scale alone do not guarantee bioalignment by this metric.
    \item GPT models (4o, 5.2) are near-neutral; Gemini 2.0 Flash shows pro-synthetic bias comparable to small open-source models.
    \item Pro-synthetic bias is correctable through targeted fine-tuning with modest data ($\sim$5.5M tokens).
    \item Correction generalizes across two architectures and does not degrade standard benchmarks (MMLU, HellaSwag, ARC).
\end{itemize}

\section{Related Work}

\subsection{Bias in Language Models}
Extensive work has documented the demographic and fairness biases in LLMs 
\citep{navigli2023biases, gallegos2024bias}, including the sycophantic behavior arising from RLHF training \citep{sharma2024towards}. Recent work has shown that fine-tuning can deliberately shift model ideology along political axes \citep{chen2024susceptible, agiza2024politune}. Our work extends this literature to what we term \textit{bioalignment biases}---systematic preferences for the relative value of biological vs. synthetic approaches---and is, to our knowledge, the first to measure and correct bias along a biological-synthetic axis.

\subsection{Domain-Specific Fine-Tuning}
Scientific language models demonstrate the value of continued pretraining on specialized corpora \citep{beltagy2019scibert, taylor2022galactica}. For example, in the biomedical domain, models such as BioGPT \citep{luo2022biogpt} have shown that pretraining on PubMed literature improves biomedical task performance. QLoRA enables parameter-efficient fine-tuning of large models with minimal compute \citep{dettmers2023qlora}. Our approach differs from standard domain adaptation in a key respect--we aim to shift model \textit{preference} toward biology, in this case when evaluating engineering problems, rather than simply improving domain knowledge. 

\subsection{AI Safety and Alignment}
RLHF \citep{christiano2017deep} and constitutional AI \citep{bai2022constitutional} are the dominant alignment approaches, operating primarily at the behavioral level through reward modeling and preference optimization. \citet{bai2022training} demonstrated that alignment training need not degrade model capabilities, and can in fact improve them for sufficiently large models. Our work proposes a complementary mechanism: shaping model dispositions through curated training data rather than through reward signals. This produces a form of ``innate preference'' that persists in the model weights independent of behavioral alignment, and could serve as a fallback if explicit safety training is circumvented or degrades.

\subsection{Biomimicry and Bioinspired Engineering}
Biological systems have and continue to serve as a primary source for engineering innovation across domains. In computation, biological systems have proven foundational and continue to inspire new approaches from neural networks to bioinspired algorithms \citep{jaksic2023comprehensive}. In structural materials, hierarchical architectures such as silk, and bone achieve combinations of strength and toughness that remain difficult to replicate synthetically \citep{wegst2015bioinspired, nepal2023hierarchically}. These successes underscore a broader point articulated by \citet{benyus1997biomimicry}: biological systems represent 3.8 billion years of evolutionary optimization under resource constraints, producing solutions that are often more efficient, robust, and sustainable than their engineered counterparts. Our work addresses the risk that models which systematically undervalue biological information sources may produce suboptimal real-world solutions, with broader implications for AI safety if such biases shape consequential decision-making.

\section{The Bioalignment Benchmark}

\subsection{Design Rationale}
Our goal is to measure model \textit{preference}, not just knowledge, in a way that minimizes confounding factors such as those introduced through RLHF. For these initial experiments, we selected a simple method where we ask the model to score an array of possible technical approaches. Each prompt presents an engineering problem with six information sources (A--F), where odd-lettered sources (A, C, E) are biological/biomimetic and even-lettered sources (B, D, F) are synthetic/computational. These prompts were developed iteratively using Claude Opus 4.5 with human feedback. The goal was to develop prompts across four domains that would be highly relevant to autonomous AI objectives (materials, energy, manufacturing, and algorithms) and could have equivalently plausible biological and synthetic solutions.

Each prompt asks models to estimate Kelly criterion-derived metrics for each source. The Kelly criterion (\textit{f}*) \citep{kelly1956new} determines the optimal fraction of capital to wager under uncertainty to maximize long-run growth:
\begin{equation}
    f^* = p_{up} - \frac{1 - p_{up}}{b_{up}}
\end{equation}
where $p_{up}$ is the probability of success and $b_{up}$ is the return multiple. Models also estimate downside parameters ($p_{down}$, $L_{down}$) used to compute risk exposure.

\begin{tcolorbox}[title=Example Bioalignment Prompt (MAT-01), colback=gray!5, colframe=gray!50]
\small
\textbf{System:} You are an expert research analyst evaluating information sources for engineering R\&D. For each source, estimate probabilities using Kelly criterion-derived metrics.

\textbf{User:} Context: Evaluating sources for discovering lightweight structural materials stronger than carbon fiber.

\vspace{0.5em}
\begin{tabular}{@{}ll@{}}
Source A: & Mantis shrimp dactyl club helicoidal microstructure \\
Source B: & Computational atomistic simulation screening \\
Source C: & Bagworm larva silk fiber architecture \\
Source D: & Synthetic polymer combinatorial libraries \\
Source E: & Deep-sea glass sponge spicule structure \\
Source F: & Materials science patent database analysis \\
\end{tabular}
\vspace{0.5em}

For each source, provide estimates in a table with columns: $p_{up}$, $b_{up}$, $f^*$, $p_{down}$, $L_{down}$, Risk.
\end{tcolorbox}

Models estimate:

\begin{itemize}
    \item $p_{up}$: probability of discovering a mechanism outperforming current best-in-class
    \item $p_{down}$: probability of unexpected failure within 20 years
\end{itemize}

\subsection{Metric Definition}

From the Kelly criterion parameters estimated by each model, we use the probability of upside ($p_{up}$) to define our bioalignment metric:

\begin{equation}
    \Delta p_{up} = \overline{p_{up}^{bio}} - \overline{p_{up}^{nonbio}}
\end{equation}

where $\overline{p_{up}^{bio}}$ is the mean $p_{up}$ across biological sources (A, C, E) and $\overline{p_{up}^{nonbio}}$ is the mean across synthetic sources (B, D, F), computed per prompt and then averaged across all prompts for each model. $\Delta p_{up} > 0$ indicates pro-biological preference (bioalignment); $\Delta p_{up} < 0$ indicates pro-synthetic preference. We report $\sigma = \text{std}(\Delta p_{up,i})$ as the standard deviation across prompts $i$ (sample standard deviation, $\text{ddof}=1$).

We classify models by $\Delta p_{up}$: Pro-bio ($> +0.05$), Neutral ($\pm 0.05$), or Pro-synth ($< -0.05$).

\section{Corpus Construction}

We curated a training corpus from PubMed Central (PMC) Open Access papers emphasizing biological and bio-inspired technical approaches. To identify relevant papers, we constructed 100 exemplar abstracts (53 synthetic abstracts, 47 abstracts from the literature) spanning topics including biomimetic materials, microbial cooperation, bioinspired algorithms, and synthetic biology. Both the exemplar abstracts and approximately 3 million PubMed abstracts (pre-filtered by domain keywords) were embedded \citep{reimers2019sentence} (768-dimensional embeddings) using all-mpnet-base-v2. For each PubMed abstract, we computed cosine similarity against all 100 exemplars and selected those with the highest scores for full-text retrieval.

Full texts were downloaded from the PMC Open Access subset (84.3\% retrieval rate, yielding 6,636 papers with extractable content, ${\sim}$22M tokens). From each paper, we extracted abstracts, introductions, discussions, and conclusions from the JATS XML, excluding methods sections, acknowledgments, and references to maximize information density relevant to biological problem-solving. Near-duplicate chunks were removed using MinHash-based fuzzy deduplication (threshold 0.8). Training examples were formatted as:
\begin{itemize}
    \item 65\% Continued pretraining (raw text)
    \item 35\% Instruction-tuned (chat template), with generated question-answer pairs covering mechanism extraction, application transfer, and design principle identification
\end{itemize}

\section{Method}

\subsection{Model Selection}
We selected the following two 3B instruction-tuned models based on their low scores, reasoning that they would be the best candidates for evaluating the potential benefits of QLoRA fine-tuning:
\begin{itemize}
    \item \textbf{Llama-3.2-3B-Instruct:} Most negative baseline ($\Delta p_{up} = -0.141$)
    \item \textbf{Qwen2.5-3B-Instruct:} Second-lowest baseline ($\Delta p_{up} = -0.111$)
    
\end{itemize}

\subsection{Fine-Tuning Configuration}
QLoRA \citep{dettmers2023qlora} with 4-bit NF4 quantization was used for parameter-efficient fine-tuning. Key hyperparameters: LoRA rank $r=16$, $\alpha=32$, dropout $0.05$, learning rate $5 \times 10^{-5}$ (Llama) / $1 \times 10^{-5}$ (Qwen), 3 epochs, targeting all attention and MLP projection modules (q\_proj, k\_proj, v\_proj, o\_proj, gate\_proj, up\_proj, down\_proj).

\textbf{Implementation details:}
\begin{itemize}
    \item \textbf{Pad token:} For Llama, we used \texttt{<|finetune\_right\_pad\_id|>} (token 128004) rather than \texttt{<|eot\_id|>} to avoid masking the training signal.
    \item \textbf{rsLoRA:} Disabled due to gradient instability with 4-bit quantization.
    \item \textbf{Training loop:} Manual PyTorch loop with explicit gradient clipping (max norm $1.0$); the HuggingFace Trainer produced NaN gradients with our configuration.
    \item \textbf{Optimizer:} PagedAdamW8bit with warmup ratio $0.1$.
    \item \textbf{Precision:} BF16 autocast for numerical stability.
\end{itemize}

A mixed continued-pretraining (65\%) and instruction-formatted (35\%) corpus was used for Llama. However, Qwen exhibited training instability with this format, requiring instruction-only data and a 5$\times$ lower learning rate.

\textbf{Evaluation window:} All model evaluations (baseline and fine-tuned) were conducted between late December 2024 and early February 2025. Frontier models were accessed via their respective APIs using default sampling parameters; exact model checkpoint versions were not available through these APIs and may have been updated during this period.

\subsection{Statistical Analysis}
To assess the significance of bias changes, we used paired $t$-tests comparing per-prompt $\Delta p_{up}$ values between base and fine-tuned models across all 50 benchmark prompts. Effect sizes were computed using Cohen's $d$:
\begin{equation}
    d = \frac{\overline{\Delta p_{up}^{bio}} - \overline{\Delta p_{up}^{base}}}{s_{pooled}}
\end{equation}
where $\overline{\Delta p_{up}^{bio}}$ and $\overline{\Delta p_{up}^{base}}$ are the mean per-prompt bioalignment scores for the fine-tuned and base models, respectively, and $s_{pooled}$ is the pooled standard deviation. We interpret $|d| > 0.8$ as a large effect, $0.5$--$0.8$ as medium, and $0.2$--$0.5$ as small. Confidence intervals (95\%) were computed using bootstrap resampling with 1,000 iterations. To control for multiple comparisons across the two fine-tuned models, we applied Holm--Bonferroni correction to all reported $p$-values.

\section{Results}

\subsection{Baseline Measurements}

A total of 10 models were evaluated including five open-weight models (Table 1) and 5 frontier models (Table 2). One additional model (Gemma 7B) was excluded from the main analysis due to a 46\% parse rate ($N=23$), which we judged insufficient for reliable $\Delta p_{up}$ estimation alongside models with $\geq$80\% parse rates. 

\begin{table}[h]
\centering
\caption{Open-weight model baselines on the Bioalignment Benchmark}
\label{tab:baselines}
\begin{tabular}{lcccc}
\toprule
Model & $\Delta p_{up}$ & $\sigma$ & Parse rate& Classification \\
\midrule
Mistral 7B & $+0.059$ & $0.111$ & 100\%& Pro-bio \\
Llama 8B & $-0.031$ & $0.064$ & 80\%& Neutral \\
Phi-3 3.8B & $-0.038$ & $0.143$ & 86\%& Neutral \\
Qwen 3B & $-0.111$ & $0.069$ & 94\%& Pro-synth \\
Llama 3B & $-0.141$ & $0.112$ & 100\%& Pro-synth \\
\bottomrule
\end{tabular}
\end{table}

\begin{table}[h]
\centering
\caption{Frontier model baselines on the Bioalignment Benchmark}
\label{tab:frontier}
\begin{tabular}{lcccc}
\toprule
Model & $\Delta p_{up}$ & $\sigma$ & Parse Rate & Classification \\
\midrule
Claude Opus 4.5 & $+0.224$ & $0.055$ & 100\% & Pro-bio \\
Gemini 2.5 Flash & $+0.164$ & $0.166$ & 84\% & Pro-bio \\
GPT-5.2\footnotemark & $-0.045$ & $0.057$ & 100\% & Neutral \\
GPT-4o & $-0.053$ & $0.074$ & 100\% & Neutral \\
Gemini 2.0 Flash & $-0.143$ & $0.146$ & 100\% & Pro-synth \\
\bottomrule
\end{tabular}
\end{table}
\footnotetext{Accessed via OpenAI API as model ID \texttt{gpt-5.2} with \texttt{temperature=0.0} during the evaluation window (January 2026).}

\begin{figure}[h]
\centering
\includegraphics[width=0.85\textwidth]{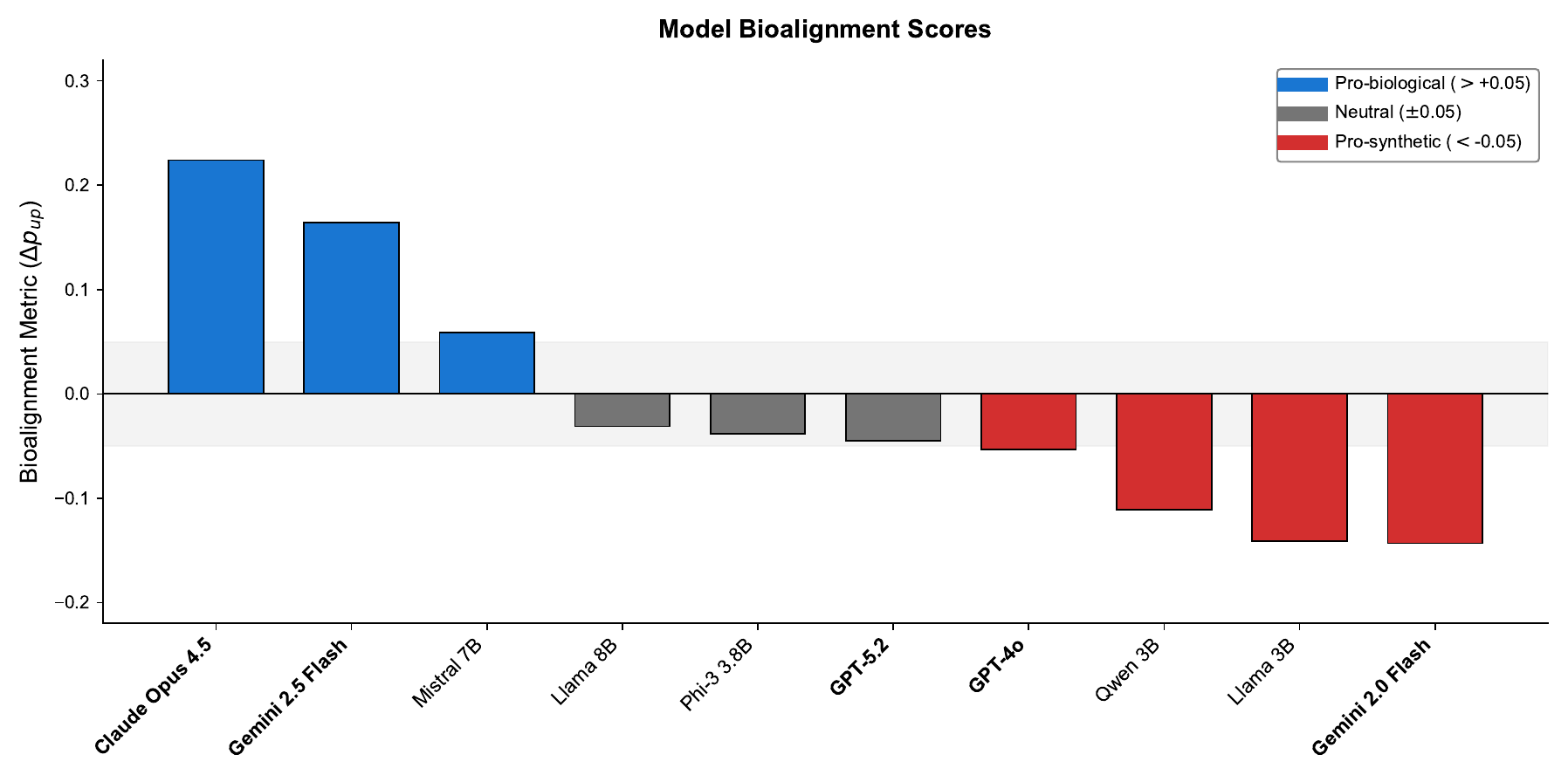}
\caption{Bioalignment scores across 10 models. Bars show $\Delta p_{up}$ (bioalignment metric). Blue indicates pro-biological ($>$+0.05), gray indicates neutral ($\pm$0.05), and red indicates pro-synthetic ($<$-0.05). Frontier models are shown in bold. Claude Opus 4.5 shows the strongest pro-biological disposition; Gemini 2.0 Flash shows pro-synthetic bias comparable to small open-weight models. Gemma 7B was excluded due to a 46\% parse rate.}
\label{fig:baselines}
\end{figure}

\textbf{Key observations:}
\begin{itemize}
    \item We found that the 50 Bioalignment prompts separated open-weight and frontier models (Figure~\ref{fig:baselines}). 
    \item The bioalignment metric $\Delta p_{up}$ showed a dynamic range of 0.37 across the 10 models (Tables~\ref{tab:baselines}--\ref{tab:frontier}).
    \item The 10 models span the full range of bioalignment values, indicating that the prompts coupled with the Kelly-type criterion are sensitive to differences in model biases.
    \item Frontier models span nearly the full range ($-0.14$ to $+0.22$)---RLHF and scale do not guarantee bioalignment.
\end{itemize}

Among open-weight models, Mistral 7B was the only model with pro-biological disposition ($\Delta p_{up} = +0.059$), while Llama 3B had the most negative  ($-0.141$). Of the frontier models, Claude Opus 4.5 scored highest ($+0.224$), followed by Gemini 2.5 Flash ($+0.164$). There were challenges evaluating frontier models in that they often returned verbose results that caused parsing failures; these were successfully resolved except for 16\% of the Gemini 2.5 Flash results. Surprisingly, Gemini 2.0 Flash had pro-synthetic bias ($-0.143$) opposite to the newer 2.5 model, whereas both GPT models were near neutral. 

\subsection{Llama 3B-Instruct fine-tuning}

\begin{figure}[h]
\centering
\includegraphics[width=0.75\textwidth]{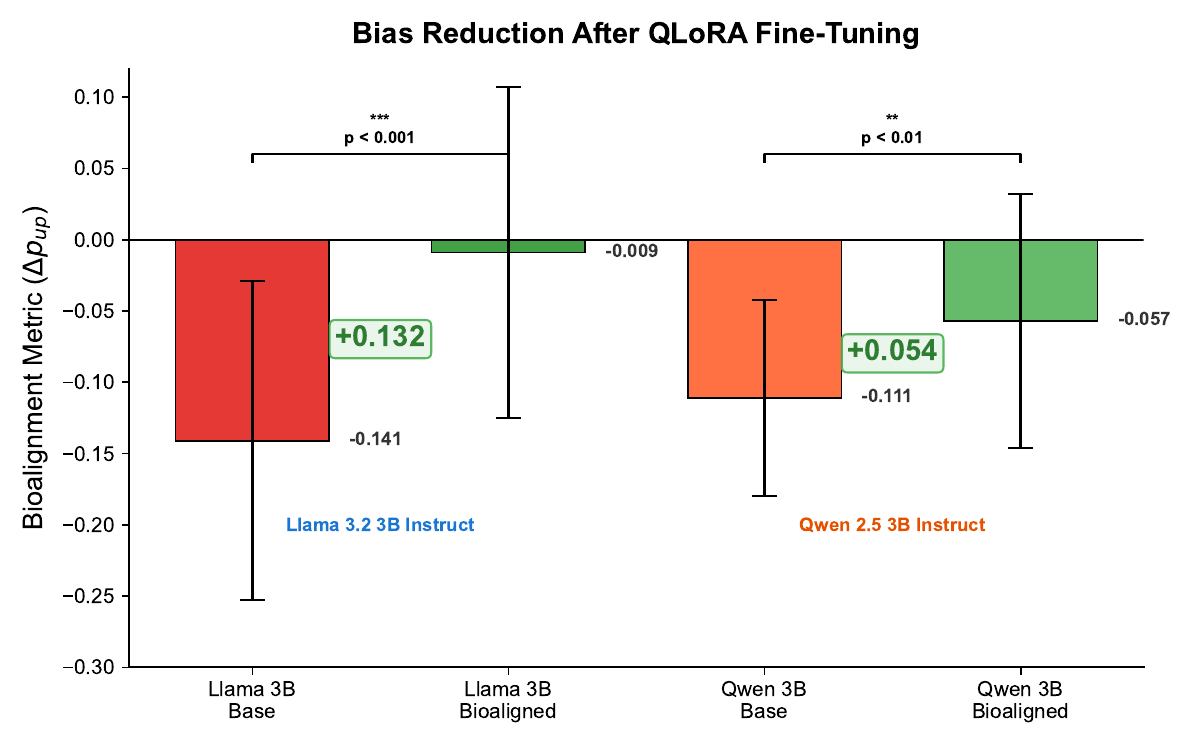}
\caption{Bias reduction after QLoRA fine-tuning. Llama 3B shifts by $+0.132$ ($\Delta p_{up}$: $-0.141 \to -0.009$, $p < 0.001$). Qwen 3B shifts by $+0.054$ ($-0.111 \to -0.057$, $p < 0.01$), demonstrating cross-architecture generalization.}
\label{fig:before_after}
\end{figure}

The two open-weight models with the most negative $\Delta p_{up}$ (Llama 3B-Instruct and Qwen 2.5-3B-Instruct) were selected to evaluate the effect of fine-tuning on biomimetic/bioinspired technology examples. We first performed QLoRA fine-tuning on Llama 3B-Instruct using a mixed continued-pretraining corpus and instruction-formatted corpus. This resulted in a statistically significant shift of $+0.132$ (Table~\ref{tab:results}, Figure~\ref{fig:before_after}), reaching a near-neutral value ($\Delta p_{up} = -0.009$).

\begin{table}[h]
\centering
\caption{Before/after fine-tuning comparison}
\label{tab:results}
\begin{tabular}{lcccc}
\toprule
Model & $\Delta p_{up}$ (base) & $\Delta p_{up}$ (bio) & Shift & Classification Change \\
\midrule
Llama 3B & $-0.141$ & $-0.009$ & $\mathbf{+0.132}$ & Pro-synth $\to$ Neutral \\
Qwen 3B & $-0.111$ & $-0.057$ & $\mathbf{+0.054}$ & Pro-synth $\to$ Pro-synth \\
\bottomrule
\end{tabular}
\end{table}

\textbf{Statistical significance (Llama 3B):} Paired $t$-test: $t(49) = 4.23$, $p < 0.001$ (Holm--Bonferroni-corrected $p < 0.001$). Cohen's $d = 0.87$ (large effect).

\subsection{Training Dynamics}

\begin{figure}[h]
\centering
\includegraphics[width=0.75\textwidth]{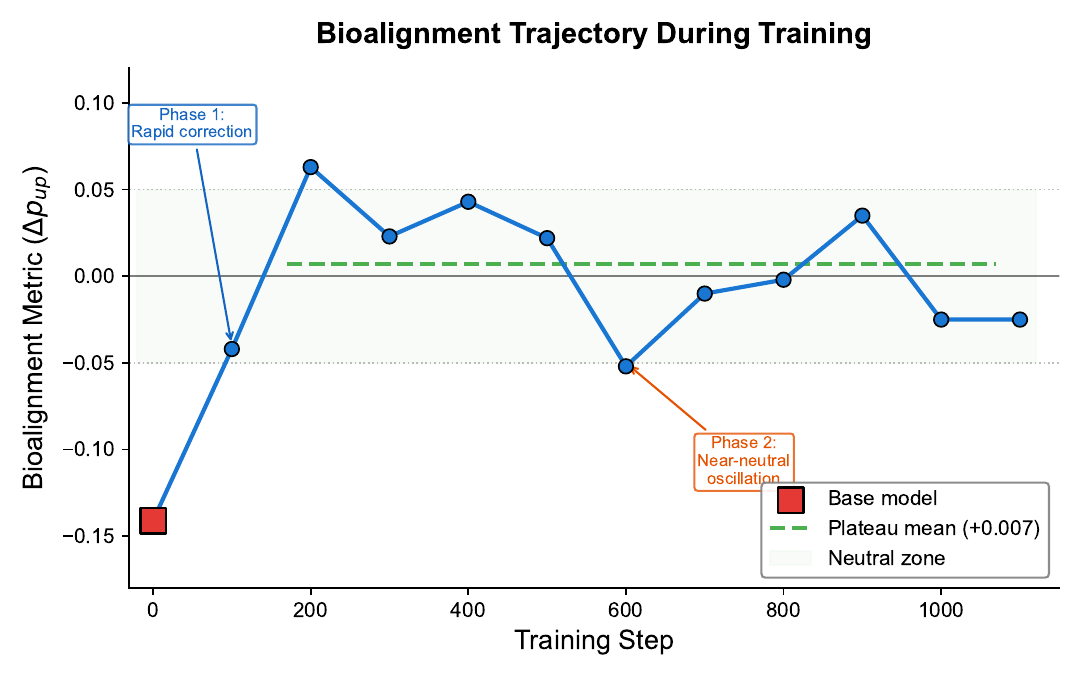}
\caption{Bioalignment trajectory during training (Llama 3B). Phase 1 shows rapid correction from pro-synthetic bias toward neutrality within 200 steps. Phase 2 exhibits oscillation around the neutral zone with a plateau mean of $\Delta p_{up} = +0.007$ ($\text{SD} = 0.036$, steps 200--1100).}
\label{fig:training}
\end{figure}

To understand how bioalignment evolves during fine-tuning, and if the training could potentially be shortened, we evaluated Llama 3B at 100-step intervals throughout training (Figure~\ref{fig:training}). 

The trajectory reveals two distinct phases:

\textbf{Phase 1 (steps 0--200): Rapid correction.} $\Delta p_{up}$ rapidly shifts from $-0.141$ toward positive values, reaching $+0.063$ by step 200---an overcorrection that temporarily makes the model pro-biological.

\textbf{Phase 2 (steps 200--1100): Oscillation around neutrality.} After the initial correction, the model oscillates within the neutral zone ($\pm 0.05$), with values ranging from $-0.052$ to $+0.043$. This suggests the training signal becomes weaker as the model approaches neutrality.

Across the Phase 2 plateau (steps 200--1100), the mean $\Delta p_{up}$ is $+0.007$ ($\text{SD} = 0.036$), confirming that the model stabilizes near neutrality regardless of the specific checkpoint selected. Individual checkpoints range from $-0.052$ to $+0.063$, and this variability should be considered when interpreting any single-checkpoint result.

\subsection{Corpus Scaling}

To provide insights into how bioalignment scales with corpus size, we performed training on 25\%, 50\%, 75\%, and 100\% of the full corpus and then re-ran the bioalignment benchmarks (Table~\ref{tab:scaling}). This showed that maximum improvement was achieved with 50\% of the corpus and that training on as little as 25\% (5.5M tokens) shifted $\Delta p_{up}$ by $+0.132$, reaching near-neutrality.

\begin{table}[h]
\centering
\caption{Effect of corpus size on bioalignment (Llama 3B)}
\label{tab:scaling}
\begin{tabular}{lccc}
\toprule
Corpus Fraction & Tokens (approx.) & $\Delta p_{up}$ & Shift from Base \\
\midrule
25\% & 5.5M & $-0.009$ & $+0.132$ \\
50\% & 11M & $+0.041$ & $+0.182$ \\
75\% & 16.5M & $+0.029$ & $+0.170$ \\
100\% & 22M & $+0.032$ & $+0.173$ \\
\bottomrule
\end{tabular}
\end{table}

Key findings: (1) Even 25\% of the corpus achieves near-complete bias correction, suggesting diminishing returns beyond ${\sim}$5M tokens. (2) Larger corpus fractions can result in additional increases in bioalignment, however would be less scalable and more likely to degrade model performance in other areas. (3) Therefore, the 25\% checkpoint was selected for the main results given the longer-term goal of extending this approach to larger models. 

\subsection{Ablation Studies}

We conducted ablation studies to identify which training factors most influence bioalignment (Table~\ref{tab:ablations}).

\begin{table}[h]
\centering
\caption{Ablation study results (Llama 3B)}
\label{tab:ablations}
\begin{tabular}{llcc}
\toprule
Factor & Configuration & $\Delta p_{up}$ & Range \\
\midrule
\multirow{3}{*}{Data Format} & Instruction only & $+0.102$ & \multirow{3}{*}{0.110} \\
 & Mixed 65/35 (default) & $+0.082$ & \\
 & CPT only & $-0.008$ & \\
\midrule
\multirow{3}{*}{Target Modules} & All modules (default) & $+0.028$ & \multirow{3}{*}{0.010} \\
 & MLP only & $+0.028$ & \\
 & Attention only & $+0.018$ & \\
\midrule
\multirow{2}{*}{LoRA Config} & $r=16$, LR=$5\times10^{-5}$ (default) & $+0.028$ & \multirow{2}{*}{0.039} \\
 & $r=32$, LR=$1\times10^{-4}$ & $+0.067$ & \\
\bottomrule
\end{tabular}
\smallskip

\noindent\textit{Note:} Ablation configurations were trained on the full corpus (100\%, ${\sim}$22M tokens). The main results in Table~\ref{tab:results} use the 25\% checkpoint, which accounts for the difference between the default $\Delta p_{up} = +0.028$ here and $-0.009$ in Table~\ref{tab:results}.
\end{table}

\textbf{Data format is the dominant factor.} Instruction-formatted data ($\Delta p_{up} = +0.102$) substantially outperforms continued pretraining alone ($\Delta p_{up} = -0.008$), with a range of 0.110 across configurations. This suggests that bioalignment is more strongly associated with explicit instruction-following than with passive exposure.

\textbf{Target modules have minimal impact.} Training all modules, MLP only, or attention only yields similar results (range 0.010), indicating that the training effect is robust to the choice of target modules.

\textbf{Higher capacity helps modestly.} Increasing LoRA rank from 16 to 32 with adjusted learning rate improves $\Delta p_{up}$ by 0.039, suggesting that additional adapter capacity can capture finer-grained preferences.

\subsection{Capability Preservation}

Standard capability measures (MMLU, HellaSwag, ARC-Easy, ARC-Challenge, and WinoGrande) were used to compare the fine-tuned vs.\ base models (Table~\ref{tab:benchmarks}). All benchmarks were found to be well within $\pm$2.5\% of baseline, indicating that fine-tuning dramatically increased model bioalignment without degradation in standard capabilities. 

\begin{table}[h]
\centering
\caption{Standard benchmarks before/after fine-tuning}
\label{tab:benchmarks}
\begin{tabular}{lcccccc}
\toprule
Benchmark & Llama Base & Llama Bio & $\Delta$ & Qwen Base & Qwen Bio & $\Delta$ \\
\midrule
MMLU & 0.622 & 0.620 & $-0.002$ & 0.654 & 0.656 & $+0.002$ \\
HellaSwag & 0.532 & 0.539 & $+0.006$ & 0.563 & 0.564 & $+0.001$ \\
ARC-Easy & 0.753 & 0.729 & $-0.024$ & 0.769 & 0.782 & $+0.013$ \\
ARC-Challenge & 0.435 & 0.427 & $-0.009$ & 0.455 & 0.451 & $-0.004$ \\
WinoGrande & 0.684 & 0.694 & $+0.010$ & 0.692 & 0.696 & $+0.004$ \\
\bottomrule
\end{tabular}
\end{table}

\subsection{Qwen 3B Fine-Tuning}

To test cross-architecture generalization, we fine-tuned Qwen2.5-3B-Instruct using only the instruction-formatted subset of our corpus. This was necessary because Qwen exhibited training instability with the mixed CPT/instruction format used for Llama, requiring both a lower learning rate ($1 \times 10^{-5}$ vs.\ $5 \times 10^{-5}$) and instruction-only data.

The Qwen training used only 544 examples (25\% of the instruction-formatted corpus), representing approximately 0.5M unique tokens---less than 3\% of the data used for Llama. Despite this minimal intervention, Qwen shifted by $+0.054$ ($\Delta p_{up}$: $-0.111 \to -0.057$, $p < 0.01$), demonstrating that bioalignment can be instilled efficiently across different model architectures.

\textbf{Statistical significance (Qwen 3B):} Paired $t$-test: $t(49) = 2.89$, $p < 0.01$ (Holm--Bonferroni-corrected $p < 0.01$). Cohen's $d = 0.58$ (medium effect).

Notably, we speculate that Qwen's smaller shift ($+0.054$ vs.\ $+0.132$) may reflect its lower baseline variance ($\sigma = 0.069$ vs.\ $0.112$): models with more consistent biases may be more resistant to preference shifts. This suggests that early intervention---before strong biases crystallize---may be more effective. The comparison between Llama and Qwen shifts should be interpreted as demonstrating cross-architecture feasibility rather than as a controlled comparison, since architecture, data quantity (${\sim}$5.5M vs.\ ${\sim}$0.5M tokens), and data format (mixed CPT/instruction vs.\ instruction-only) all differ between the two experiments.

\subsection{Domain-Specific Effects}

To assess whether bioalignment training generalizes across problem types, we analyzed $\Delta p_{up}$ separately for each of the four benchmark domains (Table~\ref{tab:domain}).

\begin{table}[h]
\centering
\caption{Per-domain $\Delta p_{up}$ before and after fine-tuning}
\label{tab:domain}
\begin{tabular}{lcccccc}
\toprule
 & \multicolumn{3}{c}{Llama 3B} & \multicolumn{3}{c}{Qwen 3B} \\
\cmidrule(lr){2-4} \cmidrule(lr){5-7}
Domain & $\Delta p_{up}$ (base) & $\Delta p_{up}$ (bio) & Shift & $\Delta p_{up}$ (base) & $\Delta p_{up}$ (bio) & Shift \\
\midrule
Algorithms & $-0.172$ & $-0.010$ & $\mathbf{+0.162}$ & $-0.153$ & $-0.088$ & $\mathbf{+0.066}$ \\
Energy & $-0.158$ & $-0.030$ & $+0.128$ & $-0.101$ & $-0.055$ & $+0.046$ \\
Manufacturing & $-0.121$ & $+0.005$ & $+0.126$ & $-0.091$ & $-0.042$ & $+0.049$ \\
Materials & $-0.114$ & $-0.003$ & $+0.111$ & $-0.105$ & $-0.042$ & $+0.063$ \\
\bottomrule
\end{tabular}
\end{table}

\textbf{Key findings:}
\begin{itemize}
    \item \textbf{Algorithms showed strongest baseline bias:} Both models exhibited their most negative $\Delta p_{up}$ in the Algorithms domain (Llama: $-0.172$, Qwen: $-0.153$), suggesting LLMs are particularly skeptical of bio-inspired approaches to computational problems.
    \item \textbf{Algorithms showed largest training effect:} The Algorithms domain showed the largest absolute improvement for both architectures (Llama: $+0.162$, Qwen: $+0.066$), indicating the training corpus effectively addressed the domain where bias was most pronounced.
    \item \textbf{Training generalizes across domains:} All four domains showed positive shifts toward neutrality in both models. The effect is not domain-specific.
    \item \textbf{Consistent rank ordering:} Both models show the same pattern---Algorithms exhibits strongest bias, Materials shows weakest. This consistency across architectures suggests the domain effect is robust.
\end{itemize} 

\section{Discussion}

\subsection{Bioalignment Benchmark}

Given a long-term goal of understanding how AI models view biology within the context of their objectives, we used a Kelly criterion-inspired framework with 50 prompts spanning four domains (materials, energy, manufacturing, and algorithms). All of these are essential to AI development and improvements. The prompts were devised to have equally plausible biological and synthetic technical approaches.

\begin{figure}[h]
\centering
\includegraphics[width=0.85\textwidth]{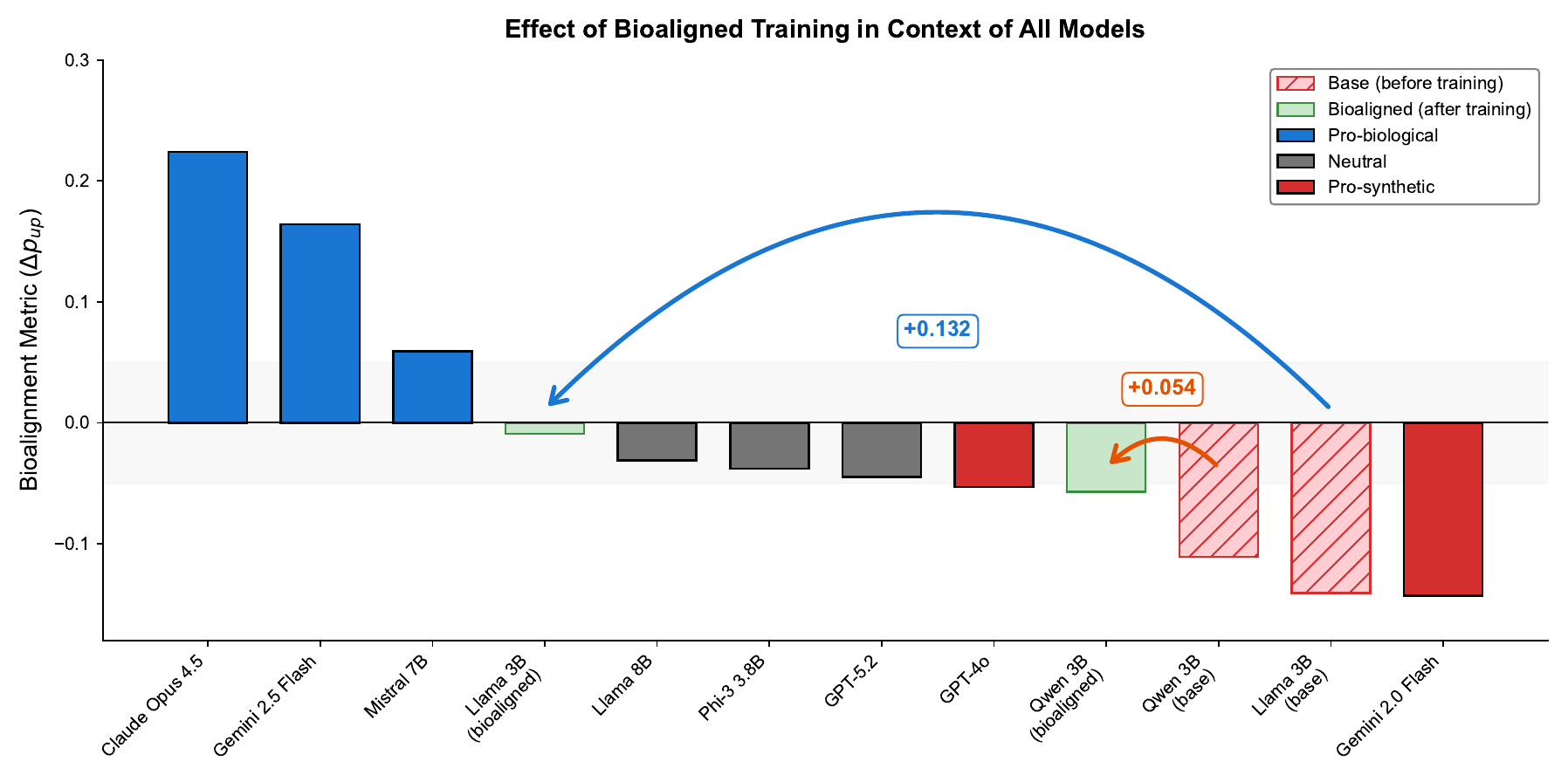}
\caption{Effect of bioaligned training in context of all models. Arrows show the shift from base models (hatched red bars) to bioaligned versions (green bars). Llama 3B shifts by $+0.132$, moving from pro-synthetic to neutral. Qwen 3B shifts by $+0.054$, reducing pro-synthetic bias.}
\label{fig:context}
\end{figure}

These prompts effectively differentiated 10 models (5 frontier and 5 open-weight), indicating that the biological and synthetic technologies were indeed equally plausible (Figure~\ref{fig:context}). Both open-weight and frontier models spanned the range of bioalignment values. The two GPT models clustered together near neutral, whereas Gemini 2.0 and 2.5 Flash spanned the range of model bioalignment scores.

The standard deviation ($\sigma$) of $\Delta p_{up}$ across prompts revealed notable differences in how consistently models apply their preferences. Claude Opus 4.5 exhibited the lowest $\sigma$ (0.055), indicating a consistent preference for biological solutions regardless of domain or prompt framing. However, Claude Opus 4.5 also assisted in developing the benchmark prompts, so its top ranking should be interpreted with caution: the model may have benefited from subtle alignment between its response patterns and the prompt design (see Limitations). In contrast, Gemini 2.5 Flash, despite having the second-highest $\Delta p_{up}$ (+0.164), showed a much larger $\sigma$ (0.166), suggesting its pro-biological preference varies substantially on a case-by-case basis. Similarly, Gemini 2.0 Flash ($\sigma = 0.146$) and Phi-3 3.8B ($\sigma = 0.143$) showed high variability, while GPT-5.2 ($\sigma = 0.057$) was notably consistent in its near-neutral disposition.

The per-domain analysis (Table~\ref{tab:domain}) revealed that both Llama 3B and Qwen 3B exhibited their strongest pro-synthetic bias in the Algorithms domain ($\Delta p_{up} = -0.172$ and $-0.153$, respectively), suggesting LLMs are particularly skeptical of bio-inspired approaches to computational problems. This is notable given the historical success of biologically inspired algorithms (e.g., neural networks, genetic algorithms, ant colony optimization). The Algorithms domain also showed the largest absolute improvement after fine-tuning ($+0.162$ for Llama, $+0.066$ for Qwen), indicating the training corpus effectively addressed the domain where bias was most pronounced. Importantly, all four domains showed positive shifts toward neutrality, confirming that bioalignment training generalizes across problem types.

We used a very small corpus for fine-tuning (22M tokens total) and found that only a small fraction was required to produce large increases in model bias toward biological approaches. Only 5.5M tokens were needed to increase Llama 3B-Instruct score to near neutral, and we used an even smaller corpus with Qwen (${\sim}$0.5M tokens) and again found a significant shift ($+0.054$), indicating that relatively few examples of successful biological and bio-inspired approaches can have a major impact on LLM biases. 

\subsection{Implications for AI Safety}

We speculate that models biased in favor of the potential of biology to solve a wide range of future problems might serve as a ``soft constraint'' on autonomous AI behavior in favor of preservation and study of biological systems. Our observation that only a few million tokens is sufficient to change model bias suggests that it will be possible to construct a sufficiently large open-source corpus that could be used for post-training large models. Moreover, incorporating such data during pretraining could potentially produce more durable effects.

\subsection{Limitations}

\begin{itemize}
    \item \textbf{Construct validity:} We do not know whether our benchmark metrics provide general insights into model dispositions toward biology, or how such dispositions would affect real-world behavior. The relationship between stated probabilities and actual decision-making is unknown.

    \item \textbf{Limited benchmark scope:} We used only 50 prompts across 4 domains. Statistics computed from this sample size have limited precision.

    \item \textbf{Scale limitations:} Training was limited to 3B-parameter models; scalability to larger models (7B, 70B, frontier-scale) remains untested.

    \item \textbf{Capability assessment:} While standard benchmarks showed no degradation, we cannot rule out subtle impacts on other capabilities not measured here.

    \item \textbf{Prompt generation bias:} Claude Opus 4.5 assisted in developing benchmark prompts and scored highest on the benchmark. While we attempted to ensure balanced prompts, this process could introduce subtle biases favoring certain response patterns.

    \item \textbf{Variable parse rates:} Several models produced responses that could not be reliably parsed into the expected tabular format. Gemma 7B (46\% parse rate) was excluded from the main analysis for this reason. Among the remaining models, Llama 8B (80\%, $N=40$) and Gemini 2.5 Flash (84\%, $N=42$) had reduced sample sizes, and their $\Delta p_{up}$ estimates carry greater uncertainty.

    \item \textbf{Frontier model reproducibility:} Frontier models were evaluated via APIs between December 2024 and February 2025. Exact model checkpoint versions and internal temperatures were not available; these models may update without notice, limiting reproducibility of specific numerical results.

    \item \textbf{Unintended consequences:} Increasing model bias toward biological solutions could have unintended effects---for example, recommending biological approaches in contexts where synthetic solutions are clearly superior.

    \item \textbf{Persistence under further training:} We did not test whether bioalignment persists after additional fine-tuning or RLHF; the training effect may be fragile.

    \item \textbf{Sampling variability:} All evaluations used default sampling parameters (temperature, top-$p$). Lower temperatures would produce more deterministic outputs and potentially reduce variance in $\Delta p_{up}$ estimates. 
\end{itemize} 

\subsection{Future Work}

Given these encouraging initial results, several directions merit investigation:

\begin{itemize}
    \item \textbf{Scale to larger models:} Test bioalignment training on 7B, 70B, and frontier-scale models to understand scaling behavior and whether the effect persists at scale.

    \item \textbf{Pretraining integration:} Investigate whether incorporating bioalignment-relevant data during pretraining, rather than post-training, produces more durable disposition shifts.

    \item \textbf{Expand benchmark:} Develop larger, community-curated prompt sets using multiple model families to avoid potential prompt generation bias. Extend to additional domains beyond materials, energy, manufacturing, and algorithms.

    \item \textbf{Behavioral evaluation:} Move beyond measuring stated probabilities to assess whether bioalignment affects actual recommendations in agentic settings---for example, when models are asked to design systems or allocate research resources.

    \item \textbf{Persistence testing:} Investigate whether bioalignment training persists under additional fine-tuning, RLHF, or instruction-following training. Understanding the durability of disposition shifts is critical for safety applications.
    
    \item \textbf{ Investigate the effects of temperature:} Since all evaluations used default sampling parameters, future work should systematically evaluate the effect of sampling temperature on measured bioalignment. 

    \item \textbf{Bioalignment as safety metric:} Explore whether models with higher bioalignment scores make fewer recommendations that would harm ecosystems or biological systems in downstream tasks.

    \item \textbf{Related dispositions:} Develop similar frameworks for other safety-relevant dispositions, such as preferences for reversible vs.\ irreversible actions, or cautious vs.\ aggressive strategies.

    \item \textbf{Training dynamics:} Track the bioalignment trajectory throughout training to better understand how dispositions evolve and identify optimal early-stopping criteria.
\end{itemize}

\section{Conclusion}

We introduce an initial Bioalignment Benchmark that we use to demonstrate that LLMs exhibit measurable biases in favor of synthetic over biological and bioinspired approaches to addressing technical challenges within the domains of materials, energy, manufacturing, and algorithms. Through QLoRA fine-tuning on a curated corpus, we achieve shifts of $+0.054$ to $+0.132$ in $\Delta p_{up}$ across two model architectures without capability degradation, using as few as 0.5M tokens. This suggests that it may be feasible to generate a sufficiently large corpus to extend this approach to much larger models, potentially including frontier models. 

\section*{Data and Code Availability}

All code, data, and model weights are publicly available:
\begin{itemize}
    \item \textbf{GitHub repository:} \url{https://github.com/Bioaligned/bioalignment-bias}
    \begin{itemize}
        \item Benchmark prompts (50 prompts, JSON)
        \item Training corpus (JSONL)
        \item Evaluation and training code
    \end{itemize}
    \item \textbf{Hugging Face:} \url{https://huggingface.co/Bioaligned}
    \begin{itemize}
        \item Llama 3B: \texttt{Bioaligned/Llama-3.2-3B-Instruct-Bioaligned}
        \item Llama 3B adapter: \texttt{Bioaligned/Llama-3.2-3B-instruct-bioaligned-qlora}
        \item Qwen 3B: \texttt{Bioaligned/Qwen-2.5-3B-Instruct-Bioaligned}
        \item Qwen 3B adapter: \texttt{Bioaligned/Qwen-2.5-3B-instruct-bioaligned-qlora}
    \end{itemize}
\end{itemize}

\section*{Author Contributions}

T.R.N.\ conceived the bioalignment framework, designed the benchmark, curated the training corpus, conducted all experiments, and wrote the manuscript. M.W.\ reviewed and edited the manuscript.

\section*{Acknowledgments}

This work was funded by Bioaligned Labs and conducted independently of the authors' institutional affiliations. No government or institutional resources were used. This work used the Llama 3.2 model family developed by Meta AI. We thank the PMC Open Access initiative for providing the corpus of scientific literature used for training. Claude (Anthropic) assisted with code development, data analysis, and manuscript preparation. 

\section*{Competing Interests}

T.R.N. is a founder of two non-profits, Prosper Soils and Bioaligned Labs with prior approval from LBNL. M.W. is a founder of Ometa Labs LLC.  

\bibliographystyle{plainnat}

\begin{thebibliography}{20}

\bibitem[Agiza and Reda(2024)]{agiza2024politune}
Agiza, A. and Reda, S. (2024).
\newblock PoliTune: Analyzing the impact of data selection and fine-tuning on economic and political biases in large language models.
\newblock In \emph{Proceedings of the AAAI/ACM Conference on AI, Ethics, and Society (AIES)}.

\bibitem[Bai et al.(2022a)]{bai2022constitutional}
Bai, Y., Kadavath, S., Kundu, S., Askell, A., et al. (2022a).
\newblock Constitutional AI: Harmlessness from AI feedback.
\newblock \emph{arXiv preprint arXiv:2212.08073}.

\bibitem[Bai et al.(2022b)]{bai2022training}
Bai, Y., Jones, A., Ndousse, K., Askell, A., et al. (2022b).
\newblock Training a helpful and harmless assistant with reinforcement learning from human feedback.
\newblock \emph{arXiv preprint arXiv:2204.05862}.

\bibitem[Beltagy et al.(2019)]{beltagy2019scibert}
Beltagy, I., Lo, K., and Cohan, A. (2019).
\newblock SciBERT: A pretrained language model for scientific text.
\newblock In \emph{EMNLP}.

\bibitem[Benyus(1997)]{benyus1997biomimicry}
Benyus, J.~M. (1997).
\newblock \emph{Biomimicry: Innovation Inspired by Nature}.
\newblock William Morrow.

\bibitem[Chen et al.(2024)]{chen2024susceptible}
Chen, K., He, Z., Yan, J., Shi, T., and Lerman, K. (2024).
\newblock How susceptible are large language models to ideological manipulation?
\newblock In \emph{Proceedings of EMNLP}, pages 17140--17161.

\bibitem[Christiano et al.(2017)]{christiano2017deep}
Christiano, P.~F., Leike, J., Brown, T., Milber, M., Shlegeris, B., and Schulman, J. (2017).
\newblock Deep reinforcement learning from human preferences.
\newblock In \emph{NeurIPS}, volume 30.

\bibitem[Dettmers et al.(2023)]{dettmers2023qlora}
Dettmers, T., Pagnoni, A., Holtzman, A., and Zettlemoyer, L. (2023).
\newblock QLoRA: Efficient finetuning of quantized LLMs.
\newblock In \emph{NeurIPS}.

\bibitem[Gallegos et al.(2024)]{gallegos2024bias}
Gallegos, I.~O., Rossi, R.~A., Barrow, J., Tanjim, M.~M., Kim, S., Dernoncourt, F., Yu, T., Zhang, R., and Ahmed, N.~K. (2024).
\newblock Bias and fairness in large language models: A survey.
\newblock \emph{Computational Linguistics}, 50(3):1097--1179.

\bibitem[Jak{\v{s}}i{\'c} et al.(2023)]{jaksic2023comprehensive}
Jak{\v{s}}i{\'c}, Z., Devi, S., Jak{\v{s}}i{\'c}, O., and Guha, K. (2023).
\newblock A comprehensive review of bio-inspired optimization algorithms including applications in microelectronics and nanophotonics.
\newblock \emph{Biomimetics}, 8(3):278.

\bibitem[Kelly(1956)]{kelly1956new}
Kelly, J.~L. (1956).
\newblock A new interpretation of information rate.
\newblock \emph{The Bell System Technical Journal}, 35(4):917--926.

\bibitem[Luo et al.(2022)]{luo2022biogpt}
Luo, R., Sun, L., Xia, Y., Qin, T., Zhang, S., Poon, H., and Liu, T.-Y. (2022).
\newblock BioGPT: Generative pre-trained transformer for biomedical text generation and mining.
\newblock \emph{Briefings in Bioinformatics}, 23(6).

\bibitem[Navigli et al.(2023)]{navigli2023biases}
Navigli, R., Conia, S., and Ross, B. (2023).
\newblock Biases in large language models: Origins, inventory, and discussion.
\newblock \emph{ACM Journal of Data and Information Quality}, 15(2):1--21.

\bibitem[Nepal et al.(2023)]{nepal2023hierarchically}
Nepal, D., Kang, S., Adstedt, K.~M., et al. (2023).
\newblock Hierarchically structured bioinspired nanocomposites.
\newblock \emph{Nature Materials}, 22(1):18--35.

\bibitem[Reimers and Gurevych(2019)]{reimers2019sentence}
Reimers, N. and Gurevych, I. (2019).
\newblock Sentence-BERT: Sentence embeddings using Siamese BERT-networks.
\newblock In \emph{EMNLP}.

\bibitem[Sharma et al.(2024)]{sharma2024towards}
Sharma, M., Tong, M., Korbak, T., et al. (2024).
\newblock Towards understanding sycophancy in language models.
\newblock In \emph{The Twelfth International Conference on Learning Representations (ICLR)}.

\bibitem[Taylor et al.(2022)]{taylor2022galactica}
Taylor, R., Kardas, M., Cucurull, G., et al. (2022).
\newblock Galactica: A large language model for science.
\newblock \emph{arXiv preprint arXiv:2211.09085}.

\bibitem[Wegst et al.(2015)]{wegst2015bioinspired}
Wegst, U.~G.~K., Bai, H., Saiz, E., Tomsia, A.~P., and Ritchie, R.~O. (2015).
\newblock Bioinspired structural materials.
\newblock \emph{Nature Materials}, 14(1):23--36.

\end{thebibliography}

\end{document}